# Predicting California Bearing Ratio with Ensemble and Neural Network Models: A Case Study from Türkiye

Abdullah Hulusi Kökçam[1], Uğur Dağdeviren[2], Talas Fikret Kurnaz[3], Alparslan Serhat Demir[1], Caner Erden[4]

[1] Department of Industrial Engineering, Faculty of Engineering, Sakarya University, Sakarya, Türkiye
[2] Department of Civil Engineering, Faculty of Engineering, Kutahya Dumlupinar University, Kutahya, Türkiye
[3] Technical Sciences Vocational School, Transportation Services, Mersin University, Mersin, Türkiye
[4] Department of Computer Engineering, Faculty of Technology, Sakarya University of Applied Sciences, Sakarya, Türkiye

**ABSTRACT**

The California Bearing Ratio (CBR) is a key geotechnical indicator used to assess the load-bearing capacity of subgrade soils, especially in transportation infrastructure and foundation design. Traditional CBR determination relies on laboratory penetration tests. Despite their accuracy, these tests are often time-consuming, costly, and can be impractical, particularly for large-scale or diverse soil profiles. Recent progress in artificial intelligence, especially machine learning (ML), has enabled data-driven approaches for modeling complex soil behavior with greater speed and precision. This study introduces a comprehensive ML framework for CBR prediction using a dataset of 382 soil samples collected from various geoclimatic regions in Türkiye. The dataset includes soil properties relevant to bearing capacity, allowing multidimensional feature representation in a supervised learning context. Twelve ML algorithms were tested, including decision tree, random forest, extra trees, gradient boosting, xgboost, k-nearest neighbors, support vector regression, multi-layer perceptron, adaboost, bagging, voting, and stacking regressors. Each model was trained, validated, and evaluated to assess its generalization and robustness. Among them, the random forest regressor performed the best, achieving strong $R^2$ scores of 0.95 (training), 0.76 (validation), and 0.83 (test). These outcomes highlight the model's powerful nonlinear mapping ability, making it a promising tool for predictive geotechnical tasks. The study supports the integration of intelligent, data-centric models in geotechnical engineering, offering an effective alternative to traditional methods and promoting digital transformation in infrastructure analysis and design.

# 1 INTRODUCTION

The California Bearing Ratio (CBR) test, first developed in 1929 by the State Highway Research Office in California to determine the bearing capacity of soils to be used in highway infrastructure, is a method used to investigate the strength of highway and airport pavements [1], [2]. CBR is defined as the ratio of the resistance of the ground at a certain penetration depth against a 49.63 mm diameter piston inserted into the ground at a speed of 1.27 mm/min to the resistance of a standard crushed stone sample at the same penetration depth [3]. The CBR value of the soil is calculated by comparing the measured loads against specific penetration values with standard values found for crushed stone. CBR tests are conducted in two ways: in the laboratory and in the field. CBR testing can be performed in the laboratory using wet CBR and dry CBR. The purpose of the wet CBR test is to determine the minimum bearing capacity at which the voids are completely filled with water. CBR is calculated based on the depth of penetration at 2.5 mm and 5 mm. The CBR value at the 2.5 mm penetration depth is normally taken into account during the design phase. If the CBR value at the 5 mm penetration depth is greater than the 2.5 mm value, the test is repeated. However, if the value is still higher in the new test, the higher CBR value is taken into account [4].

Although CBR testing provides useful information on the strength of road and airport pavements, it involves time-consuming and laborious procedures. This situation has led researchers to study the indirect methods of obtaining the CBR value. The index and compaction characteristics of soils have been frequently used in statistical approaches to determine CBR value [5]–[16]. On the other hand, with the development of computer technologies, many studies have been conducted on the estimation of CBR value with different artificial intelligence techniques. Initially, studies using different ANN architectures have been followed by research based on various machine learning techniques [17]–[25].

Given the limitations of conventional testing methods and the increasing demand for rapid, reliable geotechnical assessments, this study explores the viability of machine learning algorithms for estimating CBR values based on readily available soil parameters. This study investigates the predictive modelling of CBR values using a diverse and regionally representative dataset comprising 382 soil samples from various geographical locations across Türkiye. The dataset includes a broad spectrum of soil properties, which enables comprehensive input characterization for the machine learning algorithms. In this context, twelve distinct regression-based models were implemented and comparatively evaluated, including ensemble-based approaches (Random Forest, Gradient Boosting, AdaBoost, Bagging, Extra Trees, Voting, and Stacking), tree-based models (Decision Tree), kernel-based algorithms (Support Vector Regression), instance-based methods (K-Nearest Neighbours), and neural network architectures (Multi-Layer Perceptron). The aim is to assess the feasibility and predictive capability of these models in capturing the complex relationships between soil index properties and bearing performance, thereby offering a computational framework for efficient, scalable, and field-applicable CBR estimation.

# 2 DATA

Data used in this study consists of the CBR, Standard Proctor (SP) and the index test results of 382 soil samples. These test results are provided from the laboratory archives of the branches of the General Directorate of Türkiye Highways in different regions. The data included in the database are the CBR values, maximum dry density (MDD), optimum moisture content (OMC), liquid limit (LL), plasticity index (PI), fines content (FC), sand content (SC) and gravel content (GC). The soil samples in the database are diverse types of both fine-grained and coarse-grained samples. Statistical description of the data is given in Table 1.

Figure 1 illustrates the comparative distributions of input parameters used in the CBR prediction model, based on both training and test datasets. Each histogram displays the frequency of feature values, differentiated by data subset (Training vs. Test) to assess consistency and potential sampling bias.

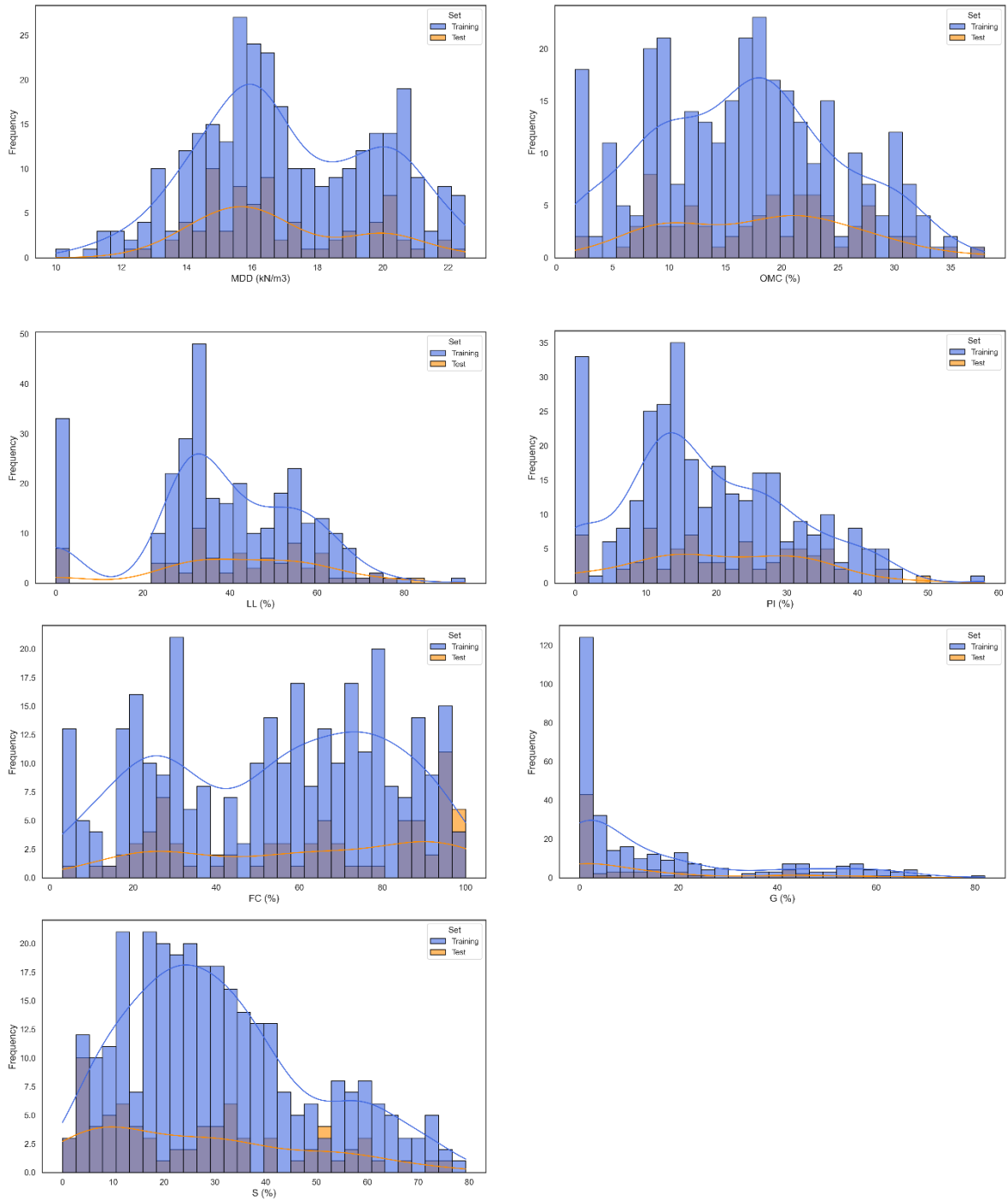

Figure 1: Distribution of features across training and test datasets

Table 1 Descriptive statistics of training and test datasets used for CBR prediction

| Set | | G (%) | S (%) | FC (%) | LL (%) | PI (%) | MDD (kN/m$^3$) | OMC (%) |
|---|---|---|---|---|---|---|---|---|
| Train | Count | 305 | 305 | 305 | 305 | 305 | 305 | 305 |
| | Mean | 15.03 | 30.74 | 54.23 | 38.02 | 18.54 | 17.12 | 16.61 |
| | Std | 19.78 | 18.05 | 27.17 | 18.27 | 11.74 | 2.69 | 8.21 |
| | Min | 0 | 1 | 3 | 0 | 0 | 10 | 1.7 |
| | 25% | 0 | 17.1 | 29.4 | 30 | 10.8 | 15.29 | 9.8 |
| | 50% | 5.2 | 28 | 58 | 36 | 16 | 16.6 | 17 |
| | 75% | 21.2 | 41 | 78 | 52 | 26.45 | 19.46 | 21.9 |
| | Max | 82 | 79.3 | 99 | 94 | 58 | 22.52 | 37 |
| Test | Count | 77 | 77 | 77 | 77 | 77 | 77 | 77 |
| | Mean | 11.93 | 26.43 | 61.64 | 41.78 | 20.97 | 16.86 | 17.97 |
| | Std | 18.17 | 20.09 | 29.47 | 19.03 | 11.77 | 2.41 | 8.07 |
| | Min | 0 | 0 | 6 | 0 | 0 | 12.5 | 1.68 |
| | 25% | 0 | 8.5 | 30.6 | 32.4 | 11.3 | 15 | 11 |
| | 50% | 2 | 24 | 65.5 | 42.4 | 20 | 16.3 | 19 |
| | 75% | 14.6 | 40 | 89 | 55.9 | 30 | 19.1 | 23 |
| | Max | 66 | 74.6 | 100 | 82.5 | 49.7 | 22.3 | 38 |

## 3   MACHINE LEARNING ALGORITHMS

Predicting natural phenomena with high accuracy has become a challenge for artificial intelligence techniques due to the inherent difficulty involved. In the last decade, Machine Learning (ML) algorithms, a component of artificial intelligence techniques, have been observed to achieve successful results in predicting these difficult phenomena. In addition to their success in predictive accuracy, machine learning algorithms can also utilize resources such as CPU and GPU more efficiently during the computational process. In this study, the following methods, which have successful applications in the literature, were used to estimate the California Bearing Ratio (CBR): XGBoost, Random Forest, Support Vector Regression (SVR), AdaBoost, Multi-layer Perceptron (MLP), K-nearest neighbors (K-NN), Bagging, Extra Trees, Voting, Gradient Boosting, Decision Tree, and Stacking, were modelled in Python and used. The study also employed the grid search method and ML techniques, hyper parametrizing their parameters to their best levels. The methodology employed in the study is illustrated in Figure 2.

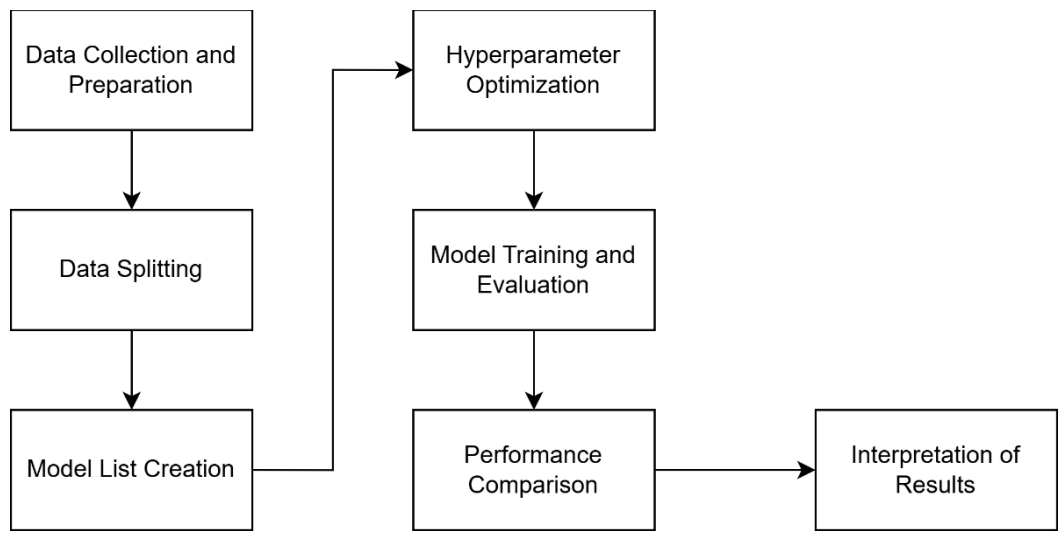

Figure 2: Flowchart of the study

## 4 RESULTS

In this study, multiple machine learning algorithms were applied to predict the CBR values based on geotechnical input parameters (GC, SC, FC, LL, PI, MDD, and OMC). The dataset consisted of 382 samples, which were randomly split into training (80%) and test (20%) subsets. This resulted in 305 training samples and 77 test samples. To ensure generalizability and prevent overfitting, hyperparameter optimization was performed using Grid Search with 5-fold cross-validation (cv=5). During this process, the training set was internally split into training and validation folds, allowing each model to be evaluated on unseen validation data across five folds. Additionally, to enhance robustness and account for variability due to random initialization, each model was trained and evaluated five times using different random state values, and the average performance metrics were reported.

Table 2 presents a comprehensive comparison of the models in terms of cross-validation (training and validation phases) and final test set performance, using $R^2$, Mean Absolute Error (MAE), and Root Mean Square Error (RMSE) as evaluation metrics.

Table 2 Comparative performance of regression models after hyperparameter tuning and cross-validation

| Model | Best Parameters | Training | | | Validation | | | Test | | |
|---|---|---|---|---|---|---|---|---|---|---|
| | | $R^2$ Avg. | MAE Avg. | RMSE Avg. | $R^2$ Avg. | MAE Avg. | RMSE Avg. | $R^2$ Avg. | MAE Avg. | RMSE Avg. |
| RandomForest | {'max_depth': 10, 'max_features': 'sqrt', 'min_samples_leaf': 1, 'min_samples_split': 5, 'n_estimators': 100} | 0.947 | 3.957 | 9.034 | 0.760 | 8.516 | 19.477 | 0.832 | 6.263 | 11.823 |
| Bagging | {'max_features': 0.7, 'max_samples': 0.9, 'n_estimators': 200} | 0.961 | 3.601 | 8.047 | 0.744 | 8.752 | 19.797 | 0.826 | 6.365 | 11.495 |
| ExtraTrees | {'max_depth': 10, 'min_samples_split': 2, 'n_estimators': 300} | 0.960 | 2.434 | 6.107 | 0.766 | 8.168 | 19.055 | 0.817 | 5.966 | 11.716 |
| Voting | N/A (No tuning parameters) | 0.994 | 1.992 | 3.315 | 0.683 | 8.662 | 21.578 | 0.812 | 6.011 | 11.611 |
| XGBoost | {'colsample_bytree': 0.7, 'gamma': 0.1, 'learning_rate': 0.01, 'max_depth': 3, 'n_estimators': 300, 'subsample': 0.7} | 0.976 | 2.702 | 5.024 | 0.708 | 8.983 | 21.521 | 0.812 | 7.173 | 12.216 |
| SVR | {'C': 1000, 'epsilon': 0.5, 'kernel': 'rbf'} | 0.799 | 6.637 | 18.692 | 0.719 | 9.248 | 21.167 | 0.780 | 6.948 | 13.766 |
| AdaBoost | {'learning_rate': 0.01, 'loss': 'exponential', 'n_estimators': 200} | 0.947 | 6.277 | 9.512 | 0.664 | 9.675 | 22.443 | 0.761 | 7.519 | 13.598 |
| KNeighbors | {'metric': 'manhattan', 'n_neighbors': 3, 'weights': 'distance'} | 0.978 | 1.120 | 2.565 | 0.615 | 9.819 | 22.609 | 0.759 | 6.850 | 14.412 |
| MLPRegressor | {'activation': 'relu', 'alpha': 0.001, 'hidden_layer_sizes': (100,), 'learning_rate': 'constant', 'solver': 'adam'} | 0.744 | 9.613 | 20.655 | 0.662 | 10.668 | 23.709 | 0.748 | 7.839 | 14.357 |
| GradientBoosting | {'learning_rate': 0.2, 'max_depth': 5, 'n_estimators': 300, 'subsample': 0.7} | 0.971 | 3.301 | 5.351 | 0.691 | 9.249 | 21.488 | 0.744 | 6.791 | 14.011 |
| DecisionTree | {'max_depth': 5, 'min_samples_leaf': 2, 'min_samples_split': 10} | 0.888 | 6.441 | 13.420 | 0.661 | 9.456 | 22.188 | 0.683 | 7.867 | 14.672 |
| Stacking | N/A (No tuning parameters) | 0.753 | 6.749 | 20.228 | 0.625 | 9.831 | 23.905 | 0.519 | 7.704 | 18.967 |

Among the evaluated models, the Random Forest Regressor achieved the highest overall performance on the final test set, with an average $R^2$ score of 0.832, MAE of 6.263, and RMSE of 11.823. However, the performance differences between Random Forest and other ensemble-based models such as Bagging, Extra Trees, and Voting Regressors were relatively small.

To further evaluate the predictive quality of the Random Forest model, two visualizations were generated. Figure 3 is a scatter plot comparing actual versus predicted CBR values on the test set, and Figure 4 is a histogram displaying the distribution of prediction errors.

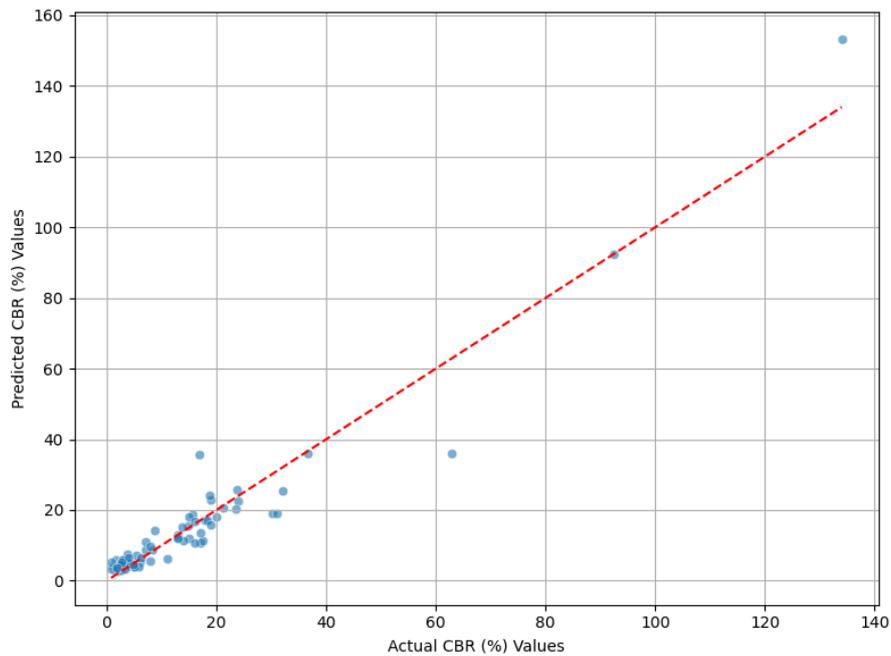

Figure 3: Actual vs. predicted CBR values using the Random Forest model on the test set

As shown in Figure 3, most of the data points are located close to the diagonal (red dashed) line, which represents a perfect prediction (y = x). While slight deviations exist—particularly at higher CBR values—predictions are generally well aligned with actual values, confirming the model's accuracy. A few outliers, especially at the extreme ends of the CBR range, can be observed; these may be attributed to sparsity in those regions of the training data or to the model's limitations in extrapolation.

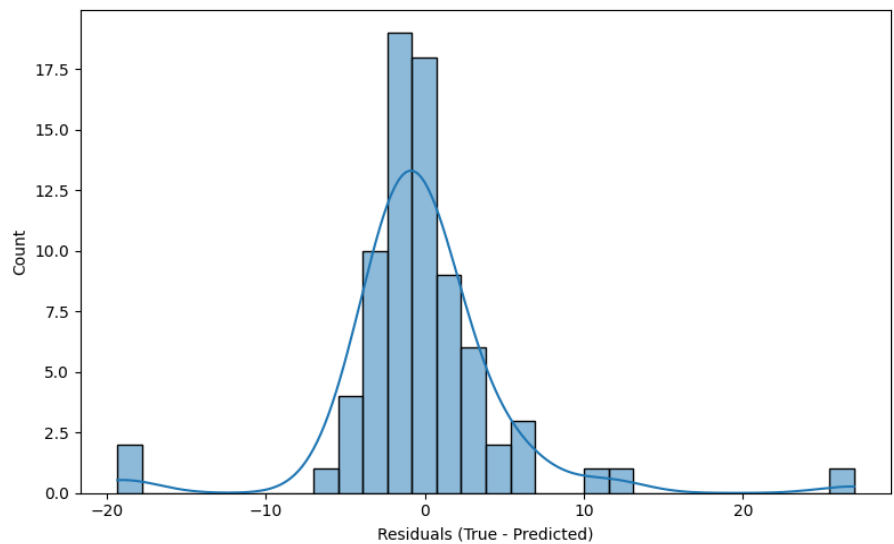

Figure 4: Distribution of prediction errors (Actual - Predicted) for the Random Forest model on the test set

The error distribution in Figure 4 follows a near-normal distribution centered around zero, indicating that the model's errors are unbiased. Most of the residuals fall within the range of -10 to +10, with a slight skew toward negative errors, suggesting a tendency for slight overestimation in certain cases. Extreme residuals are rare, which supports the model's robustness.

Figure 5 presents a comparison between the actual and predicted CBR values for the test set samples. The Random Forest regression model effectively predicts CBR values, closely matching actual measurements in the test set. It successfully captures trends, including peaks, troughs, and outliers, demonstrating strong generalization. The predicted and actual values nearly overlap, indicating the model has effectively learned data relationships and offers robust performance on unseen samples.

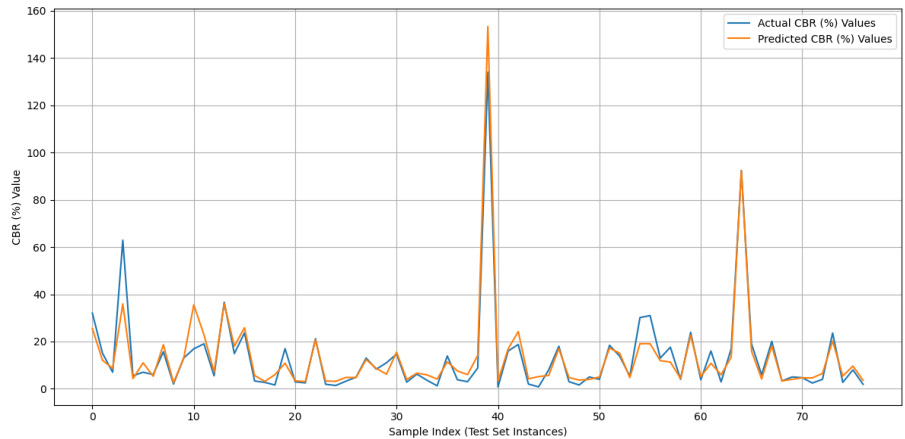

Figure 5: Comparison of actual and predicted CBR values for the test set samples.

In summary, the Random Forest Regressor emerged as the most accurate and stable model for predicting CBR values among the eleven tested algorithms. The combination of high R² scores, low prediction errors, and consistent performance across validation and test sets underscores its suitability for this geotechnical application.

## 5   CONCLUSION

This study has demonstrated the potential of machine learning algorithms as a viable alternative to traditional laboratory-based procedures for estimating the CBR, an essential indicator in geotechnical engineering. By leveraging a comprehensive dataset of 382 soil samples collected across diverse regions in Türkiye, the research explored multiple data-driven approaches—ranging from ensemble methods to neural networks—to model the complex relationships between soil properties and bearing capacity. The comparative analysis of twelve regression-based models revealed that Random Forest consistently outperformed other algorithms in terms of predictive accuracy and generalization. This outcome affirms the robustness of ensemble-based learning techniques in capturing nonlinear patterns inherent in geotechnical datasets. From a practical standpoint, the integration of machine learning into geotechnical workflows offers significant advantages: reduced testing time, lower costs, and enhanced scalability in infrastructure diagnostics. These findings not only support the adoption of artificial intelligence for indirect CBR estimation but also contribute to the ongoing transformation of civil engineering practices toward intelligent and automated decision-support systems.